\documentclass[lettersize, journal]{IEEEtran}

\usepackage[final]{changes}
\usepackage{changes}
\usepackage{amsmath, amsfonts}
\usepackage{algorithmic}
\usepackage{algorithm}
\usepackage{array}
\usepackage[caption=false, font=normalsize, labelfont=sf, textfont=sf]{subfig}
\usepackage{textcomp}
\usepackage{stfloats}
\usepackage{url}
\usepackage{verbatim}
\usepackage{graphicx}
\usepackage{cite}
\usepackage{amssymb}
\usepackage{graphics} 
\usepackage{epsfig}
\usepackage{booktabs}
\usepackage{multirow}
\usepackage[utf8]{inputenc}
\usepackage{url}
\usepackage{booktabs}
\usepackage{amssymb}
\usepackage{bbding}
\usepackage{pifont}
\usepackage{wasysym}
\usepackage{utfsym}
\usepackage{fontawesome}
\usepackage{booktabs}
\usepackage{multirow}
\usepackage{amsmath}
\usepackage{hyperref}
\usepackage{mathrsfs}
\usepackage[numbers, sort&compress]{natbib}
\usepackage{txfonts}
\begin{document}

\title{SDS-Net: Shallow-Deep Synergism-detection Network for infrared small target detection}

%\author{Taoran~Yue,~Xiaojin~Lu,~Jiaxi~Cai,~Yuanping~Chen,~Shibing~Chu~
\author{
	Taoran~Yue,~\IEEEmembership{Graduate student Member, IEEE},
	Xiaojin~Lu,
	Jiaxi~Cai,
	Yuanping~Chen,
	Shibing~Chu$^{*}$,~\IEEEmembership{Society Member, IEEE}
\thanks{Shibing~Chu$^{*}$ denotes Corresponding author, email:c@ujs.edu.cn}
\thanks{All the authors are with School of Physics and Electronic Engineering, Jiangsu University, Zhenjiang, China.}
\thanks{This work gratefully acknowledges the National Natural Science Foundation of China (No. 11904137, 12074150 and 12174157).}

%\thanks{This work gratefully acknowledges the National Natural Science Foundation of China (No. 11904137, 12074150 and 12174157).}

All the authors are with School of Physics and Electronic Engineering, Jiangsu University, Zhenjiang, China. %(email: 2222326019@stmail.ujs.edu.cn; 2222326032@stmail.ujs.edu.cn; 2222326007@stmail.ujs.edu.cn;chenyp@ujs.edu.cn;c@ujs.edu.cn)

}

\markboth{Journal of \LaTeX\ Class Files,~Vol.~14, No.~8, May~2025}%
{Shell \MakeLowercase{et al.}: A Sample Article Using IEEEtran.cls for IEEE Journals}

\maketitle
\begin{abstract}
%\textcolor{blue}\emph{S}patial-channel \emph{C}ross \emph{T}ransformer \emph{Net}work~(SCTransNet)
 
Current CNN-based infrared small target detection (IRSTD) methods generally overlook the heterogeneity between shallow and deep features, leading to inefficient collaboration between shallow fine-grained structural information and deep high-level semantic representations. Additionally, the dependency relationships and fusion mechanisms across different feature hierarchies lack systematic modeling, which fails to fully exploit the complementarity of multilevel features. These limitations hinder IRSTD performance while incurring substantial computational costs. To address these challenges, this paper proposes a shallow-deep synergistic detection network (SDS-Net) that efficiently models multilevel feature representations to increase both the detection accuracy and computational efficiency in IRSTD tasks. SDS-Net introduces a dual-branch architecture that separately models the structural characteristics and semantic properties of features, effectively preserving shallow spatial details while capturing deep semantic representations, thereby achieving high-precision detection with significantly improved inference speed. Furthermore, the network incorporates an adaptive feature fusion module to dynamically model cross-layer feature correlations, enhancing overall feature collaboration and representation capability. Comprehensive experiments on three public datasets (NUAA-SIRST, NUDT-SIRST, and IRSTD-1K) demonstrate that SDS-Net outperforms state-of-the-art IRSTD methods while maintaining low computational complexity and high inference efficiency, showing superior detection performance and broad application prospects. Our code will be made public at \url{https://github.com/PhysiLearn/SDS-Net}.

\end{abstract}

\section{Introduction}
\label{sec:IN}
%~\cite ~\ref{fig1}
\IEEEPARstart{I}{nfrared} dim and small target detection (IRSTD) is crucial for applications such as early warning, traffic management, aerospace, and maritime rescue ~\cite{1,2,3,4,5,6,7}. Unlike visible-light imaging, infrared systems capture thermal radiation, providing all-weather detection with strong anti-interference ability and environmental penetration ~\cite{8}. Early IRSTD methods primarily focused on model-driven approaches, including spatial filtering via Top-Hat transforms and Butterworth filters to suppress background noise ~\cite{9}, bioinspired vision models with multichannel Gabor filters to enhance salient features ~\cite{10,11}, and low-rank/sparse decomposition techniques such as RPCA for background isolation ~\cite{12,13}. Although effective in specific scenarios, these methods rely on hand-crafted heuristics, lack semantic understanding, and struggle with dynamic backgrounds, limiting their performance in real-world conditions.
\begin{figure*}[ht]
	\centering
	\includegraphics[width=.99\textwidth]{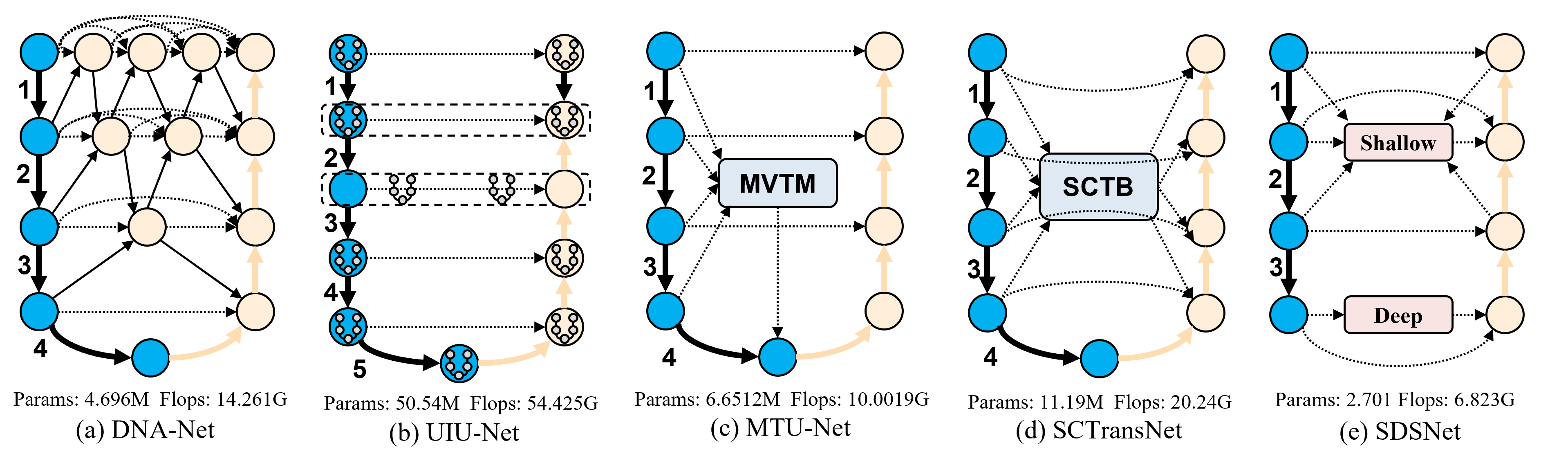}
	%\vspace{-1mm}
	\caption{Framework and visualization diagram of representative IRSTD methods.}
	%\vspace{-3mm}
	\label{u-net}
\end{figure*}
\newline \indent With the rise of deep learning and GPU computing, CNN-based methods have significantly advanced infrared target detection. However, bounding box-based localization continues to face challenges with small target shapes, boundaries, and background noise, leading to performance degradation. Pixel-level semantic segmentation has thus become the dominant approach in IRSTDs, offering finer target extraction and better robustness in complex environments ~\cite{14}.
\newline \indent Infrared small targets in remote sensing typically present three challenges: (1) extremely small size (3×3 to 9×9 pixels, ~0.01\% of the image area ~\cite{15}); (2) high shape variability influenced by ambient conditions; and (3) low contrast against cluttered backgrounds, making them vulnerable to noise ~\cite{16}.
\newline \indent UNet-based models have effectively addressed these challenges. ACM-Net ~\cite{17} uses asymmetric context modulation for multilevel fusion; DNA-Net ~\cite{18} introduces nested interactions for better representation; and UIU-Net ~\cite{19} enhances multiscale modeling via cross-attention. Despite their success, these models focus on local features and lack global context modeling, which is critical when targets resemble background noise.
\newline \indent To address this issue, hybrid architectures that integrate transformers with U-Net structures have been proposed ~\cite{20,21}. As illustrated in Fig.~\ref{u-net}(c)(d), transformer blocks are inserted and refine skip connections are refined to enhance semantic consistency. While improving target-background separability through global-local feature fusion, their quadratic complexity poses significant computational challenges.
\newline \indent Through an in-depth analysis of current mainstream methods, we identify three limitations:
\textbf{(1) The loss of fine-grained details and the increased computational cost from repeated downsampling.} Owing to the inherently small size of infrared targets, their feature representations are limited. Repeated downsampling further degrades spatial resolution, eroding critical structural details and weakening cross-layer feature interactions. Additionally, high-resolution inputs significantly increase the computational burden.
\textbf{(2) The uniform processing of shallow and deep features limits their complementarity.} Visualization reveals that shallow layers capture local textures vital for localization, whereas deep layers encode semantic context for discrimination. Treating them uniformly neglects their distinct roles, thereby reducing detection effectiveness. Targeted processing is thus essential to exploit their complementary strengths.
\textbf{(3) Coarse fusion without dependency modeling impairs feature integration.} Simple concatenation or weighting fails to account for the semantic-structural gap between shallow and deep features, leading to ineffective fusion. In small target detection, this can cause misalignment and localization errors. Modeling bidirectional dependencies is key to achieving coordinated structure-semantic integration.
\begin{figure*}[ht]
	\centering
	\includegraphics[width=.99\textwidth]{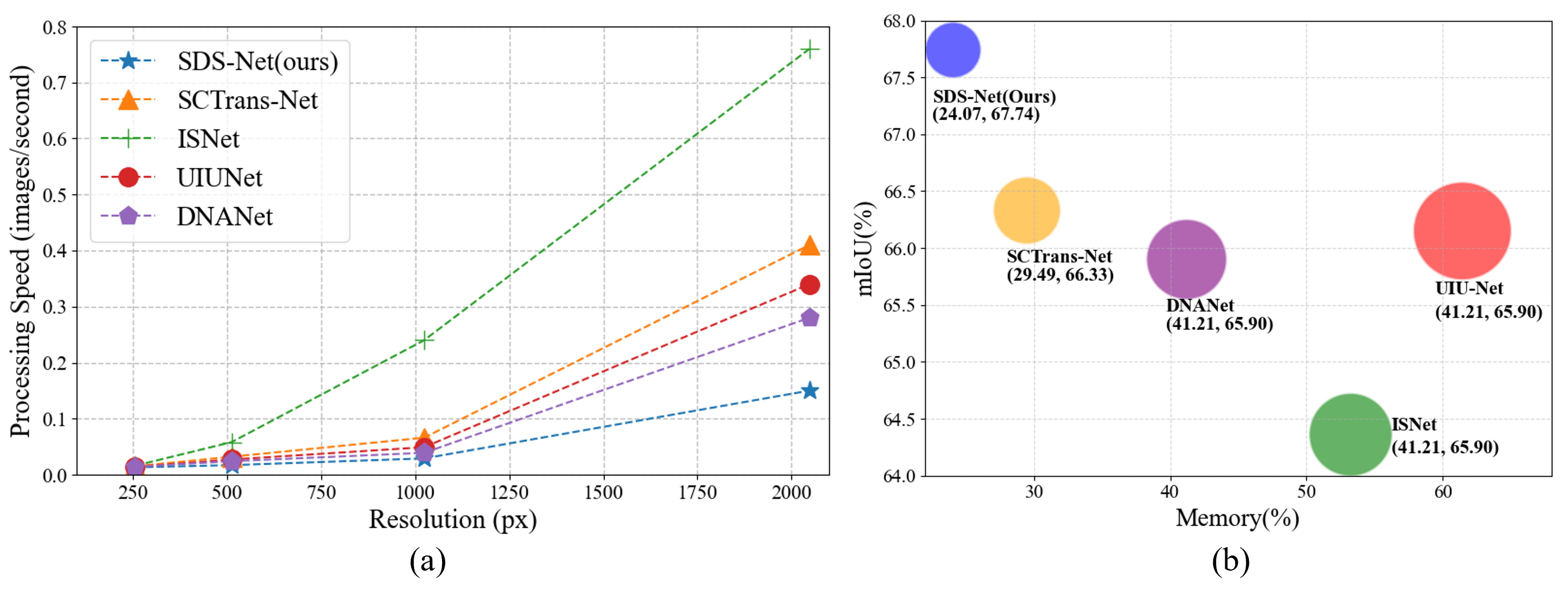}
	%\vspace{-1mm}
	\caption{(a). Our SDS-Net is more computationally and memory efficient than the present SOTA methods. (b) The overall efficiency comparison on images of resolution 512 × 512, where larger bubbles denote higher GPU memory usage.}
	%\vspace{-3mm}
	\label{rcp}
\end{figure*}
\newline \indent This paper proposes the Shallow-Deep Synergistic Detection Network (SDS-Net) to address limitations in accuracy, memory usage, and inference speed commonly seen in IRSTDs. Inspired by bilateral structures in lightweight segmentation models ~\cite{22,23}, SDS-Net adopts a dual-branch architecture that integrates shallow and deep pathways. Unlike traditional bilateral designs that use identical inputs, SDS-Net introduces a heterogeneous input strategy, assigning scale-specific inputs to each branch on the basis of their respective roles. This design enables the branches to generate complementary features, improving the overall detection performance. Additionally, an adaptive feature fusion mechanism is further introduced to model interlayer dependencies while avoiding redundant computations. This enhances feature interaction efficiency and representational capacity. As shown in Fig.~\ref{rcp}(a), SDS-Net achieves faster inference speeds than existing methods do, with increasing benefits at higher resolutions. Fig.~\ref{rcp}(b) shows a 2.5× reduction in GPU memory usage compared with that of UIU-Net, saving approximately 37.4\% of the memory per image. Overall, SDS-Net delivers a strong balance of accuracy and efficiency, outperforming current mainstream approaches. The main contributions of this work are summarized as follows:
\newline \indent 1) Shallow-Deep Synergistic Detection Framework: A dual-path architecture designed for the IRSTD, where heterogeneous inputs to the shallow and deep branches improve feature complementarity and detection accuracy.
\newline \indent 2)	Targeted Dual-Branch Design: The shallow branch incorporates multiscale spatial cross-attention (MSCA) and multidimensional dynamic fusion attention (MDFA) modules for fine-grained detail enhancement, whereas the deep branch uses multiscale spatial self-attention (MSSA) and MDFA to balance between semantic abstraction and detail preservation.
\newline \indent 3)	Adaptive Deep-Shallow Fusion (ADSF): A learnable fusion mechanism that dynamically adjusts feature contributions, enabling effective hierarchical integration and improving discriminability between targets and the background.
\newline \indent 4)	Extensive Experimental Validation: Compared with start-of-the-art methods comprehensive evaluations on three public IRSTD datasets demonstrate significant gains in detection accuracy, inference speed, and memory efficiency.
\newline \indent The paper is organized as follows: Section~\ref{Sec2} reviews related work. Section~\ref{sec3} details the SDS-Net framework. Section~\ref{sec4} presents the experimental results and analysis. Section~\ref{sec5} concludes the paper and discusses future directions.
\section{Related work}
\label{Sec2}
\subsection{Unet-based and multi-scale feature extraction IRSTD methods}
\label{Sec2-A}
The U-Net architecture, proposed by Ronneberger et al. ~\cite{24}, features a symmetric encoder-decoder structure with skip connections, enabling accurate spatial localization and inspiring subsequent work in IRSTD. Dai et al. ~\cite{17} introduced the asymmetric context modulation (ACM) method, which leverages bidirectional feature propagation to validate the effectiveness of cross-layer fusion. This strategy has since been widely adopted in IRSTD models. ALC-Net ~\cite{25} uses the bottom-up local attention modulation (BLAM) module and a parameter-free local contrast module, which enables long-range contextual interactions beyond conventional convolutional limits. DNA-Net ~\cite{18} introduces a densely nested interaction module (DNIM) to facilitate progressive fusion between low- and high-level features. AGPC-Net ~\cite{26} designs the attention-guided context block (AGCB) to model local correlations within feature patches, followed by global context attention (GCA) to capture long-range dependencies. ISNet ~\cite{27} advanced this approach by proposing the two-way oscillatory attention aggregation (TOAA) module for cross-scale fusion and the texture feature decoder (TFD) for edge enhancement, along with the release of the IRSTD-1K dataset to promote future research. To address class imbalance, Sun et al. ~\cite{28} developed the receptive field and direction-induced attention network (RDIAN), which leverages multiscale receptive fields to increase target feature diversity.
\newline \indent Despite these advancements, UNet-based methods are limited by their local receptive fields, making them less effective at modeling long-range dependencies, which is crucial for distinguishing small targets from complex infrared backgrounds.

\subsection{UNet-based methods combined with transformer IRSTD methods}
\label{Sec2-B}
Attention mechanisms, as the core component of transformers, compute inter-element dependencies through dot-product operations among queries, keys, and values. This approach enables effective global context modeling and parallel computation, providing innovative solutions for the IRSTD. For example, Qi et al. ~\cite{29} proposed FTC-Net, a hybrid UNet–transformer framework where the U-Net branch captures local details while the transformer branch leverages hierarchical self-attention to model global dependencies and suppress background interference. Tong et al. ~\cite{30} introduced GA-JL, which incorporates a dual-branch detection head for multiresolution feature extraction and a guided attention module combining spatial and channel attention to enhance discriminative features. In MSAFFNet  ~\cite{31}, a dual attention module (DAM) was designed to refine both shallow spatial details and deep semantic features. Huang et al.  ~\cite{32} presented FDDBA-Net, which applies frequency-domain decoupling with learnable masks to extract target-specific spectra, along with a bidirectional attention module for local-global feature interaction. Nian et al.  ~\cite{33} proposed LCAGNet, which introduces a multiscale attentive local contrast module (MALC) and a collaborative attention fusion module (MCAF) to integrate detail-rich low-level features with abstract high-level semantics.
\newline \indent However, these existing methods treat multilevel features uniformly in transformer modules, ignoring their varying semantic granularity and representation capabilities. Tailored processing of these hierarchical features can better exploit their complementary strengths, enhancing both efficiency and detection performance.

\subsection{Deep-Shallow Framework in Vision Tasks}
\label{Sec2-C}
Recent advances in hierarchical feature modeling emphasize the importance of heterogeneous multilevel processing for fully exploiting deep and shallow features ~\cite{34,35}. For example, Yang et al. ~\cite{36} proposed SDCL, a transformer-based shallow-deep collaborative framework for unsupervised visible-infrared person reidentification, which improves cross-modal alignment by leveraging low-level structural cues and high-level semantics. Cheng et al. ~\cite{37} introduced DMF²Net for heterogeneous remote sensing change detection, combining central difference convolution with grayscale-gradient priors to enhance spatial detail retention. Guo et al. ~\cite{38} developed ISDNet for ultrahigh-resolution image segmentation, integrating shallow and deep branches through a relation-aware fusion module to increase feature discriminability and segmentation accuracy.
\newline \indent These studies demonstrate that the shallow-deep framework in vision tasks effectively improves both representation quality and computational efficiency. Inspired by these approaches, we propose a shallow-deep synergistic detection mechanism for IRSTDs that enhances detail sensitivity in the early layers and semantic abstraction in the deeper layers, ensuring a robust separation between targets and the background.

%\vspace{-1mm}
\section{Method}
%\vspace{-1mm}
\label{sec3}
This section provides a detailed overview of the proposed SDS-Net for infrared small target detection in detail. Section III-A describes the overall network architecture. Section III-B explores the design of the dual-branch module design, specifying the configurations and roles of the shallow and deep branches. Section III-C presents the ADSF module, which integrates hierarchical features to enhance target representation.

\subsection{Overall pipeline}
\label{sec3-A}
\begin{figure*}[ht]
	\centering
	\includegraphics[width=.99\textwidth]{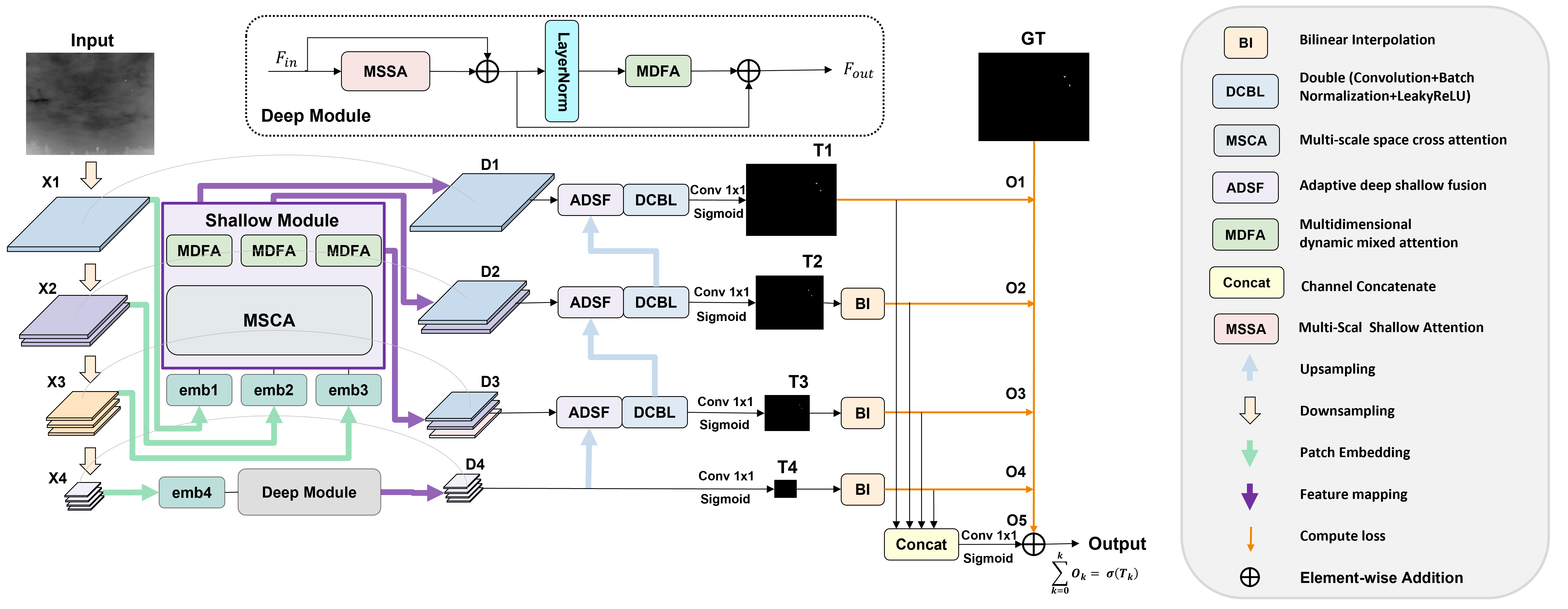}
	%\vspace{-1mm}
	\caption{Overview of SDS-Net and deep modules for infrared small target detection. SDS-Net adopts a U-shaped deep and shallow layer framework and conducts targeted processing through deep and shallow layer modules. The above is the processing procedure diagram of the deep module.}
	%\vspace{-3mm}
	\label{overall}
\end{figure*}

The overall architecture of SDS-Net is illustrated in Fig.~\ref{overall}. Its core design employs a three-stage feature fusion strategy to progressively integrate shallow-level detail features and deep-level semantic information, thereby significantly enhancing the detection performance for small targets in infrared images. The implementation workflow is detailed as follows:
\newline \indent First, the input image undergoes multiscale feature extraction through a backbone network, generating four hierarchical feature maps denoted as $X_i \in \mathbb{R}^{C_i \times \frac{H}{i} \times \frac{W}{i}}, \; i \in \{1, 2, 3, 4\}$, where the channel dimensions are $C_1 = 32$, $C_2 = 64$, $C_3 = 128$, and $C_4 = 128$. To unify the spatial resolution of feature maps, scale adjustment is performed via convolutional layers with varying kernel sizes and strides (specifically $P$, $P/2$, $P/4$, and $P/4$). This process produces uniformly sized feature representations $E_i \in \mathbb{R}^{C_i \times \frac{H}{16} \times \frac{W}{16}}, \; i \in \{1, 2, 3, 4\}$. Here, $E_1$, $E_2$, and $E_3$ are routed to the shallow feature branch, whereas $E_4$ is directed to the deep semantic branch.
\newline \indent The Feature Mapping (FM) module, a critical component, consists of bilinear interpolation, convolutional operations, batch normalization, and ReLU activation. This module enhances and fuses high-resolution shallow features with deep semantic features, ultimately outputting reconstructed features $D_i \in \mathbb{R}^{C_i \times \frac{H}{i} \times \frac{W}{i}}, \; i \in \{1, 2, 3, 4\}$.The entire feature processing pipeline can be formally expressed as:
\begin{equation}
	\begin{aligned}
		D_i &= \mathrm{X}_i + \mathrm{FM}_i\big(\mathrm{Shallow\_Module}(\mathrm{E}_1, \mathrm{E}_2, \mathrm{E}_3)\big),\quad i \in \{1, 2, 3\} \\
		D_4 &= \mathrm{X}_4 + \mathrm{FM}_4\big(\mathrm{Deep\_Module}(\mathrm{E}_4)\big)
	\end{aligned}
\end{equation}
\newline \indent During the fusion phase, an Adaptive Deep-Shallow Fusion (ASDF) module is introduced to enhance feature interaction capabilities. This module fuses upsampled deep features or fused features from the previous layer with current shallow features. The fusion results are further processed through the DCBL decoding module, which consists of two sets of $3 \times 3$ convolutions, batch normalization, and ReLU activation functions. This hierarchical process progressively generates fused features $F_i$, with the specific calculations as follows:
\begin{equation}
	\begin{aligned}
		F_3 &= \mathrm{DCBL}\big(\mathrm{ASDF}(D_3, \mathrm{upsample}(D_4))\big) \\
		F_i &= \mathrm{DCBL}\big(\mathrm{ASDF}(D_i, \mathrm{upsample}(F_{i+1}))\big), \quad i \in \{1, 2\}
	\end{aligned}
\end{equation}
\newline \indent For enhanced training efficiency and prediction accuracy, SDS-Net incorporates a multilevel supervision mechanism. Mask prediction is applied to the output features $F_i$ $(i=1,2,3)$ at each level and the deep feature $D_4$ through $1 \times 1$ convolutional layers and sigmoid activation, producing multilevel candidate maps $T_i$. The formulation is expressed as follows:
\begin{equation}
	\begin{aligned}
		T_i &= \sigma(f_{1 \times 1}(F_i)), \quad i \in \{1, 2, 3\}
		\\
		T_4 &= \sigma(f_{1 \times 1}(D_4))
	\end{aligned}
\end{equation}
\newline \indent After upsampling $T_2$, $T_3$, and $T_4$ to the original image resolution, all feature maps are concatenated and processed through a $1 \times 1$ convolutional layer followed by sigmoid activation ($\sigma$) to generate the final fused prediction map $T_5$.
\begin{equation}
	T_5 = \sigma\big(f_{1 \times 1}(\mathrm{Concat}(T_1, \mathcal{B}(T_2), \mathcal{B}(T_3), \mathcal{B}(T_4)))\big)
\end{equation}
Herein, $B$ denotes the bilinear interpolation upsampling operation, and $\mathrm{Concat}(\cdot)$ represents the channel-wise concatenation. To achieve multiscale supervised optimization, the binary cross-entropy (BCE) loss is computed between all the prediction maps and the ground truth (GT) masks, with a weighted summation of these terms constituting the final loss function.
\begin{equation}
	O_1 = \mathcal{L}_{\mathrm{BCE}}(T_1, G), \quad O_5 = \mathcal{L}_{\mathrm{BCE}}(T_5, G)
\end{equation}
\begin{equation}
	O_i = \mathcal{L}_{\mathrm{BCE}}(\mathcal{B}(T_i), G), \quad i \in \{2, 3,4\}
\end{equation}
\begin{equation}
	\mathcal{L} = \sum_{i=1}^{5} O_i
\end{equation}
\subsection{Shallow-Deep Synergistic Detection}
\label{sec3-B}
\begin{figure*}[ht]
	\centering
	\includegraphics[width=.99\textwidth]{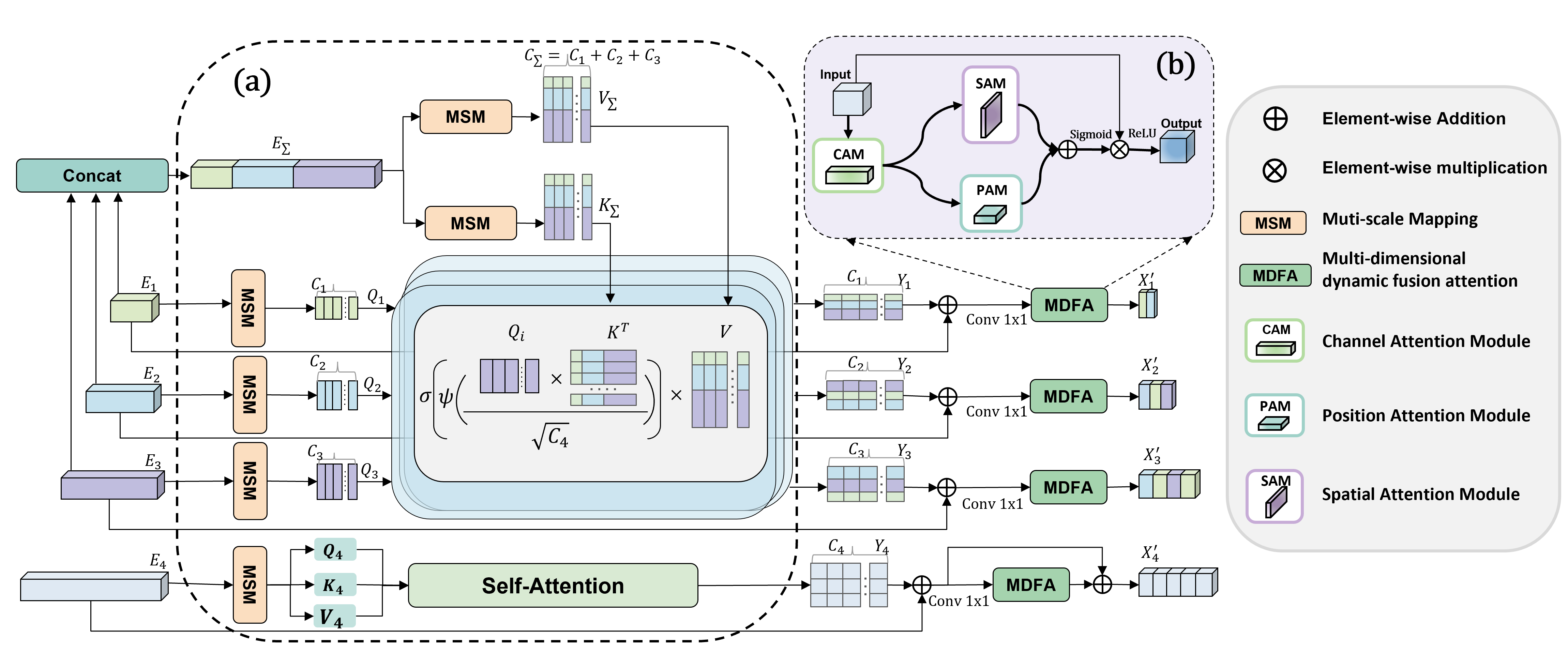}
	%\vspace{-1mm}
	\caption{Shallow-Deep Synergistic Framework. (a) MSCA captures spatial dependencies via shallow feature interactions. MSSA increases deep advanced semantic expression (b) MDFA integrates multiscale features to enhance context and geometry.}
	%\vspace{-3mm}
	\label{M2STM}
\end{figure*}

Recently, methods such as UIU-Net, DNA-Net, and ABC-Net ~\cite{39} have made significant strides in suppressing complex backgrounds and enhancing small target features in infrared imagery. However, these approaches primarily emphasize deep semantic extraction while overlooking shallow high-resolution details, which often resulting in the attenuation or loss of small targets during feature propagation. Recognizing the complementary roles of shallow spatial details and deep semantic features, this paper introduces a collaborative shallow-deep detection framework designed to fully leverage multilevel representations fully and improve detection performance.
\newline \indent To retain fine-grained target information, we propose a multi-scale spatial cross-attention (MSCA) module that extends traditional cross-attention with multilevel semantic modeling, enhancing sensitivity to detailed features. Simultaneously, a multi-scale self-attention (MSSA) module enriches contextual semantics in deep layers. The fused features are further refined by a multidimensional dynamic feature augmentation (MDFA) module, which adaptively enhances discriminative capability through dynamic weighting. Detailed implementations of these modules are provided in the following sections.
\begin{figure}[ht]
	\centering
	\includegraphics[width=.49\textwidth]{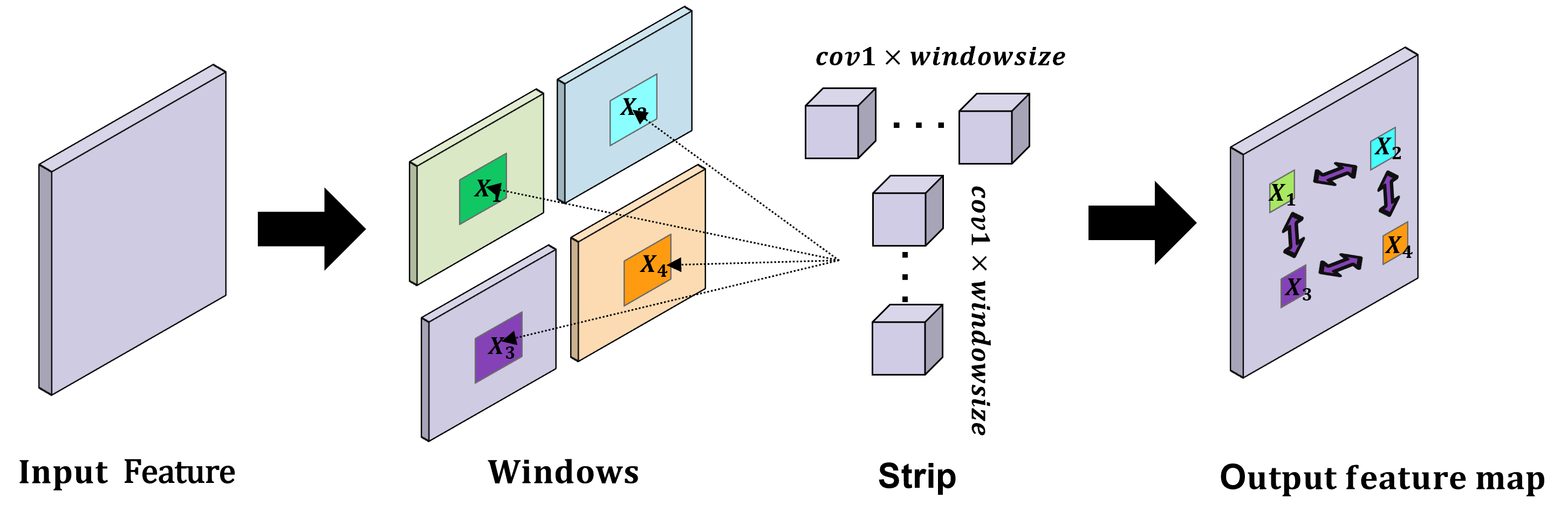}
	%\vspace{-1mm}
	\caption{Vertical Stripe Convolution for Enhanced Window Interaction in Feature Maps.}
	\vspace{-7mm}
	\label{MSM}
\end{figure}
1) Multiscale Spatial Cross-Attention and Self-Attention (MSCA and MSSA). Inspired by the Swin Transformer’s ~\cite{15} ability in local-global feature modeling, we introduce vertical strip convolutions to enhance long-range dependency modeling across windows. This approach employs convolutional groups with kernel sizes of $(1 \times x)$ and $(x \times 1)$ to facilitate inter-window information interaction, effectively improving long-range spatial dependency representations for precisely capturing sparse and faint small target features in infrared images.
\newline \indent As illustrated in Fig.~\ref{MSM}, feature connections from $x_1$ to both $x_2$ and $x_3$ are achieved through convolutions of length equal to the window size. Additionally, dependencies between $x_2$ and $x_3$ to $x_4$ are established, with each pixel possessing information from other pixels within its respective window, thereby facilitating interaction between windows.
\newline \indent As shown in Fig.~\ref{M2STM}(a), we first extract multiple shallow features $E_1, E_2, E_3$, where $E_i \in \mathbb{R}^{C_i \times \frac{H}{16} \times \frac{W}{16}}$, $(i=1,2,3)$. These features are concatenated to obtain the integrated shallow feature representation $E_{\Sigma} \in \mathbb{R}^{C_{\Sigma} \times \frac{H}{16} \times \frac{W}{16}}$, where $C_{\Sigma} = C_1 + C_2 + C_3$. To fully exploit spatial features across scales, all features undergo multiscale mapping (MSM), a module comprising multiple vertical strip convolution groups designed to establish representation capabilities under varying receptive fields. The formal implementation is as follows:
\begin{equation}
	\small
	\centering
	f_{\mathrm{MAM}}(X) = \delta_{1 \times 1} \big( \mathrm{OC}_7(\mathrm{LN}(X)) + \mathrm{OC}_{11}(\mathrm{LN}(X)) + \mathrm{OC}_{21}(\mathrm{LN}(X)) \big)
	\label{eq:MAM}
\end{equation}
\newline \indent Here, $\mathrm{OC}_i$ denotes strip convolutions with kernel sizes of $1 \times k$ and $k \times 1$, $\mathrm{LN}(\cdot)$ represents layer normalization, and $f_{\mathrm{MAM}}(\cdot)$ signifies the multi-scale mapping operation. Building on MSM, we further construct the query ($Q$), key ($K$), and value ($V$) matrices required for the attention mechanism.
\begin{equation}
	K_{\Sigma} = f_{\mathrm{MAM}}(E_{\Sigma}), \quad V_{\Sigma} = f_{\mathrm{MAM}}(E_{\Sigma})
	\label{eq:KV_sum}
\end{equation}
\newline \indent The shallow features are mapped to query vectors:
\begin{equation}
	Q_i = f_{\mathrm{MAM}}(E_i), \quad i \in \{1, 2, 3\}
	\label{eq:Q_shallow}
\end{equation}
\newline \indent The deep feature $E_4$ undergoes MSM to generate its corresponding query and key-value pairs:
\begin{equation}
	Q_4, K_4, V_4 = f_{\mathrm{MAM}}(E_4)
	\label{eq:QKV_deep}
\end{equation}
\newline \indent The shallow features are subsequently fed into the MSCA module, whereas the deep features are processed by the MSSA module. The specific attention computations are formulated as follows:
\begin{equation}
	\begin{aligned}
		Y_i &= \mathrm{OC}_1 \left( \mathrm{Softmax}\left\{ \mathcal{I} \left( \frac{Q_i K_{\Sigma}^\top}{\lambda} \right) \right\} V_{\Sigma} \right), \quad i \in \{1, 2, 3\} \\
		Y_4 &= \mathrm{OC}_1 \left( \mathrm{Softmax} \left\{ \mathcal{I} \left( \frac{Q_4 K_4^\top}{\lambda} \right) \right\} V_4 \right)
	\end{aligned}
\end{equation}
\newline \indent where $\mathcal{I}(\cdot)$ denotes the instance normalization operation~\cite{ref61}, $\lambda$ is an optional temperature factor, $Y_i \in \mathbb{R}^{C_i \times h \times w}$ represents the output of MSCA, and $Y_4 \in \mathbb{R}^{C_4 \times h \times w}$ corresponds to the output of MSSA.
\newline \indent 2) Although CBAM-based ~\cite{40} attention mechanisms have shown strong performance in large-scale object detection by enhancing channel responses and semantic modeling, their ability to localize faint infrared small targets remains limited due to insufficient positional information modeling. Existing methods~\cite{41,42} often overlook the interplay among positional, spatial, and channel dimensions, leading to reduced responsiveness in low-contrast conditions.
\newline \indent To address this challenge, we propose the MDFA module, which not only refines the integration strategy of CBAM but also incorporates a positional attention module (PAM) to form a unified positional-spatial-channel interaction framework. By jointly leveraging spatial, channel, and positional cues, this design enhances positional sensitivity and enables adaptive feature fusion through dynamic weighting, thereby significantly improving detection performance in complex infrared scenarios.
\newline \indent As illustrated in Fig.~\ref{M2STM}(b), for the input feature $Y_i \in \mathbb{R}^{C_i \times h \times w}$ and its associated residual branch $E_i$, we first aggregate them via summation before processing through a channel attention module (CAM):
\begin{equation}
	F_{c_i} = \sigma\big( \mathrm{MLP}(\mathrm{GAP}(Y_i + E_i)) + \mathrm{MLP}(\mathrm{GMP}(Y_i + E_i)) \big)
	\label{eq:Fc}
\end{equation}
\newline \indent where $\mathrm{GAP}(\cdot)$ and $\mathrm{GMP}(\cdot)$ represent global average pooling and max pooling operations, respectively. $\mathrm{MLP}(\cdot)$ consists of a two-layer fully-connected structure (first layer: $C/8$ channels, second layer: restored to $C$ channels). Subsequently, $F_{c_i}$ is fed into the spatial attention module (SAM) to obtain the spatially enhanced feature $F_{s_i}$:
\begin{equation}
	F_{s_i} = f^{7 \times 7}\big( \mathrm{DL}(\mathrm{GMP}(F_{c_i})); \mathrm{DL}(\mathrm{GMP}(F_{c_i})) \big)
	\label{eq:Fs}
\end{equation}
\newline \indent Here, $f^{7 \times 7}$ indicates a $7 \times 7$ convolutional kernel with dilation rate 4, $\mathrm{DL}(\cdot)$ denotes a dense layer, and ‘;’ signifies channel-wise concatenation. The spatial feature $F_{s_i}$ subsequently enters the Positional Attention Module (PAM), generating three reshaped matrices $B, Z, D \in \mathbb{R}^{H \times W \times C}$ via convolution:
\begin{equation}
	s_{xy} = \frac{\exp(B_x \cdot Z_y)}{\sum_{x=1}^N \exp(B_x \cdot Z_y)}
	\label{eq:sxy}
\end{equation}
\begin{equation}
	F_{P_i} = \alpha \sum_{y=1}^N (s_{xy} \cdot D_x) + F_{c_i}
	\label{eq:Fp}
\end{equation}
The positional output $F_{P_i}$ and spatial output $F_{s_i}$ are adaptively fused through sigmoid-gated multiplication with $Y_i$, followed by ReLU activation:
\begin{equation}
	X_i' = \mathrm{ReLU}\big( \sigma(F_{P_i} + F_{s_i}) \cdot Y_i \big)
	\label{eq:final_fusion}
\end{equation}
\newline \indent By enabling dynamic interactions across spatial, positional, and channel dimensions, MDFA facilitates multigranularity feature integration and adaptive attention modulation. This design effectively preserves the global context while significantly enhancing the detectability and robustness of small infrared targets in complex backgrounds.

\subsection{Adaptive Deep-Shallow Fusion Module}
\label{sec3-C}
Deep–shallow feature fusion is essential for infrared small target detection. Shallow features retain fine spatial details but are sensitive to noise, while deep features provide semantic abstraction but often miss small targets owing to low resolution.
\newline \indent The proposed ADSF module learns cross-level dependencies and fusion weights, enabling effective complementarity between detailed shallow features and semantically rich deep features. This enhances both localization and recognition, improving detection robustness under complex backgrounds.
\begin{figure*}[t!]
   \centering
    \includegraphics[width=.99\textwidth]{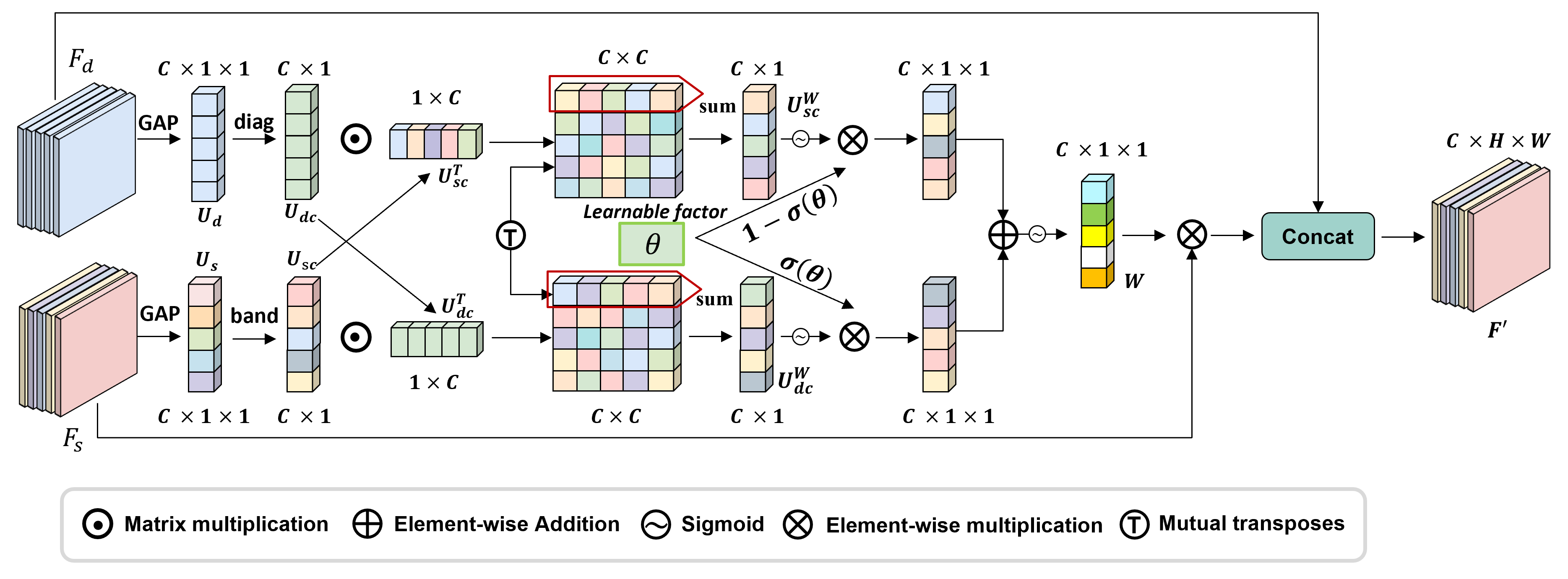}
    \caption{Adaptive deep-shallow feature fusion.}
    \label{ADSF}
\end{figure*}
\newline \indent As illustrated in Fig.~\ref{ADSF}, the input $F_s$ is regarded as the shallow feature map, whereas $F_d$ denotes either the upsampled deep feature or the hybrid feature map from the preceding stage. First, global average pooling (GAP) is applied to both shallow and deep features to compress them into channel-wise descriptors:
\begin{equation}
	U_s = \mathrm{GAP}(F_s), \quad U_d = \mathrm{GAP}(F_d)
	\label{eq:gap}
\end{equation}
\newline \indent The GAP operation compresses the feature map from $C \times H \times W$ to $C \times 1 \times 1$, aggregating the global response information per channel. To model local channel interactions in shallow features and global channel dependencies in deep features while maintaining parameter efficiency, we introduce a band matrix $B = [b_1, b_2, \dots, b_k]$ and a diagonal matrix $D = [d_1, d_2, \dots, d_c]$ for transforming $U_s$ (shallow) and $U_d$ (deep), respectively. The shallow and deep channel encodings $U_{sc}$ and $U_{dc}$ are defined as:
\begin{equation}
	U_{dc} = \sum_{i=1}^{k} U \circ b_i, \quad
	U_{sc} = \sum_{i=1}^{c} U \circ d_i
	\label{eq:encoding}
\end{equation}
\newline \indent where $\circ$ denotes elementwise multiplication, $k$ is the local bandwidth (bandpass channels), and $c$ is the channel dimension. These operations, implemented via 1D/2D convolutions, refine features through localized band filtering ($U_{dc}$) and channel-wise diagonal modulation ($U_{sc}$). The compatibility between shallow and deep channels is quantified by the correlation matrix:
\begin{equation}
	M = U_{sc} \circ U_{dc}^\top
	\label{eq:correlation}
\end{equation}

To adaptively regulate information flow, we derive weight vectors from $M$ and dynamically fuse features under the guidance of a learnable factor $\Theta$:
\begin{equation}
	U_{sc}^w = \sum_j^c M_{i,j}, \quad i \in \{1, 2, \dots, c\}
	\label{eq:usc_weight}
\end{equation}
\begin{equation}
	U_{dc}^w = \sum_j^c M^\top_{i,j}, \quad i \in \{1, 2, \dots, c\}
	\label{eq:udc_weight}
\end{equation}
\begin{equation}
	W = \sigma\left( \sigma(\Theta) \times \sigma(U_{sc}^w) + (1 - \sigma(\Theta)) \times \sigma(U_{dc}^w) \right)
	\label{eq:fusion_weight}
\end{equation}
\newline \indent Here, $W$ represents the fusion weights, dynamically balancing shallow edge-texture emphasis ($U_{sc}^w$) and deep semantic-target prioritization ($U_{dc}^w$) via sigmoid-activated gating $\sigma$. The fused feature $F'$ is generated via channel-weighted integration:
\begin{equation}
	F' = \mathrm{Concat}(F_d, W \otimes F_s)
	\label{eq:fusion}
\end{equation}
\newline \indent This strategy minimizes interlayer redundancy while amplifying cross-hierarchy semantics, achieving superior adaptability and discriminative capability for infrared small targets in cluttered environments.
\section{Experiments and Analysis}
\label{sec4}
\subsection{Datasets prepare} 
\label{sec4-A}
In our experiments, we utilized three public datasets: NUAA-SIRST ~\cite{17} (427 images, $256 \times 256$ pixels), NUDT-SIRST ~\cite{18} (1,327 images, $256 \times 256$ pixels), and IRSTD-1K ~\cite{27} (1,001 images, $512 \times 512$ pixels). All the datasets adopt an 8:2 training--test split. Notably, nearly 50\% of the samples contain targets occupying $\leq 0.2\%$ of the image area. Furthermore, the datasets encompass complex remote sensing scenarios (sky, ground, buildings, and sea) with diverse target types. This configuration rigorously reflects the challenging nature of infrared small target detection.

\subsection{Experimental details} 
\label{sec4-B}
All experiments were conducted on a Linux system with an Intel Xeon Platinum 8481 CPU, 24\,GB RAM, and an NVIDIA GeForce GTX 4090 D GPU (CUDA 11.8), using the PyTorch framework. The {Adam optimizer ($\beta=0.9$) was employed with MSE loss, an initial learning rate of $1\times10^{-3}$ (minimum $1\times10^{-5}$), and a total of 1,000 epochs with cosine annealing for learning rate scheduling. 
Batch sizes were set to 4 for NUAA-SIRST/NUDT-SIRST and 16 for IRSTD-1K, with a non-maximum suppression (NMS) IoU threshold of 0.5. SCTransNet was adopted as the baseline model, and no pre-trained weights were used.
\newline \indent To evaluate SDS-Net, we compared it with 12 state-of-the-art IRSTD methods (Top-Hat ~\cite{9}, WSLCM ~\cite{43}, IPI ~\cite{44}, ACM ~\cite{17}, ALCNet ~\cite{25}, RDIAN ~\cite{28}, ISTDU-Net ~\cite{45}, IAANet ~\cite{46}, DNA-Net ~\cite{18}, UIU-Net ~\cite{19}, SCTransNet ~\cite{21}) on three benchmarks. For fairness, all methods were retrained with the same dataset and followed the original protocols with fixed thresholds.

\subsection{Evaluation metrics} 
\label{sec4-C}
The evaluation employs pixel-level metrics (IoU, nIoU) and detection metrics (Precision ($P_d$), Recall ($F_a$), F-measure), supplemented by ROC analysis. Key metrics are defined as follows:  
\begin{enumerate}
	\item IoU:This metric measures the contour alignment between the predictions and the ground truth:  
	\begin{equation}
		\text{IoU} = \frac{\sum_{i=1}^{N} \text{TP}[i]}{\sum_{i=1}^{N} \left( \text{T}[i] + \text{P}[i] - \text{TP}[i] \right)}
	\end{equation}
	
	\item nIoU: Normalized per-sample IoU averaging:  
	\begin{equation}
		\text{nIoU} = \frac{1}{N} \sum_{i=1}^{N} \frac{\text{TP}[i]}{\text{T}[i] + \text{P}[i] - \text{TP}[i]}
	\end{equation}
	\item F1-Score: Harmonic mean of Precision (Prec) and Recall (Rec):  
	\begin{equation}
		F = \frac{2 \times \text{Prec} \times \text{Rec}}{\text{Prec} + \text{Rec}}
	\end{equation}	
	\item $P_d$: Ratio of correctly predicted targets ($N_{\text{pred}}$) to total targets ($N_{\text{all}}$):  
	\begin{equation}
		P_d = \frac{N_{\text{pred}}}{N_{\text{all}}}
	\end{equation}	
	\item False Alarm Rate ($F_a$): Ratio of falsely detected pixels ($N_{\text{false}}$) to total pixels ($P_{\text{all}}$):  
	\begin{equation}
		F_a = \frac{N_{\text{false}}}{P_{\text{all}}}
	\end{equation}
\end{enumerate}

To address class imbalance, ROC analysis dynamically evaluates the $P_d$-$F_a$ trade-off across varying thresholds, where a higher $P_d$ at an equivalent $F_a$ indicates superior performance.

\subsection{Compared with other SOTA methods} 
\label{sec4-D}
Comparative experiments with state-of-the-art methods (Table~\ref{tab1}) demonstrate the superiority of our approach. The proposed SDS-Net achieves optimal performance across pixel-level metrics (mIoU, nIoU, and F-measure) and detection metrics ($P_d$, $F_a$), surpassing existing infrared small target detection methods in quantitative evaluations.
\begin{table*}[t]
	\renewcommand{\arraystretch}{1.3}
	\caption{Comparisons with SOTA methods on NUAA-SIRST, NUDT-SIRST and IRSTD-1K in $IoU(\%)$, $nIoU(\%)$, $F$-$measure(\%)$, $P_d(\%)$, $F_a({10}^{-6})$.}
	\label{tab1}
	\setlength\tabcolsep{1.4mm}
	\begin{tabular}{c|ccccc|ccccc|ccccc} \hline
		\multirow{3}{*}{Method} & \multicolumn{5}{c|}{NUAA-SIRST~\cite{7}}                                           & \multicolumn{5}{c|}{NUDT-SIRST~\cite{8}}                                           & \multicolumn{5}{c}{IRSTD-1K~\cite{13}}                                           \\ \cline{2-16} 
		
		& mIoU      & nIoU     & {F-measure}       & Pd              & Fa             & mIoU      & nIoU     & {F-measure}       & Pd              & Fa             & mIoU      & nIoU     & {F-measure}       & Pd              & Fa             \\\hline
		Top-Hat~\cite{9}                 & 7.143     & 18.27      & {14.63}    & 79.84          & 1012          & 20.72     &  28.98    & {33.52}    & 78.41           & 166.7          & 10.06     & 7.438    &{16.02}    & 75.11           & 1432         \\
		WSLCM~\cite{43}              & 1.158     & 6.835      & {4.812}    & 77.95           & 5446         & 2.283     &  3.865   & {5.987}    & 56.82           & 1309           & 3.452     & 0.678    &{2.125}    & 72.44           & 6619          \\
		IPI~\cite{44}                 & 25.67     &  50.17    & {43.65}    & 84.63           & 16.67         & 17.76     &  15.42    & {26.94}    & 74.49          & 41.23          & 27.92     & 20.46    &{35.68}    & 81.37           & 16.18          \\
		ACM~\cite{17}                  & 70.78     & 71.85    & {78.55}    & 95.41           & 13.14          & 61.12     & 64.40    & {75.47}    & 94.18           & 34.61          & 59.23     & 57.03    &{68.29}    & 93.27           & 65.28          \\
		ALCNet~\cite{25}                  & 70.83     & 71.05    & 89.07    & 94.30           & 36.15          & 64.74     & 67.20    & {77.69}    & 94.18           & 34.61          & 60.60     & 57.14    & {71.89}    & 92.98           & 58.80          \\		
		RDIAN~\cite{28}                  & 71.99     & 76.90    & 82.55   & 93.54           & 43.29          & 76.28     & 79.14    & {85.74}    & 95.77           & 34.56          & 56.45     & 59.72    & {68.89}    & 88.55           & 26.63          \\
		ISTDU~\cite{45}              & 77.52     & 79.73    & \textbf{91.85}    & 96.58           & 14.54          & 89.55     & 90.48    & {93.85}    & 97.67           & 13.44          & 64.36     & 63.86    & 73.49   & \underline{93.60}           & 53.10          \\
		IAANet~\cite{46}   & 75.31    & 74.65    & 84.78   & 96.53    & 22.70 
		& 90.22     &  92.04   & 95.01    & 97.26    & 8.32   
		& 65.25     & 64.77    & 75.36    & 93.15    & 14.20  \\		
		ISNet~\cite{27}   & 74.38     & 76.82    & 83.76    & 95.06   & 32.76 
		& 85.57    & 82.91    & 95.78    & 97.78    & 6.34   
		& 64.36    & 65.07    &  76.15   & 90.24      & 31.56          \\
		DNA-Net~\cite{18}               & 76.14     & \underline{81.28}    & 82.55    & 97.24           & 26.11         & 88.19     & 88.58    & {94.82}    & \underline{98.83}           & 9.00           & 65.90     & 65.38    & {76.42}    & 90.91           & 12.24          \\
		UIU-Net~\cite{19}              & \underline{80.26}     & 80.88    & 85.63    & \underline{98.16}           & \underline{8.874}          & 93.48     & 93.89    & {\underline{96.99}}    & 98.31           & 7.79           & 66.15     & \underline{66.66}    &  78.24    & \textbf{93.93}           & 22.07          \\
		SCTransNet ~\cite{21}                & 79.84     & 80.24    & 87.32    & 96.95          & 13.92          & \underline{94.32}     & \underline{94.12}    & 96.95    & 98.62           & \textbf{4.29}           & \underline{66.33}     & 66.42    &  \underline{79.84}    & 91.27           & \underline{10.74}          \\ 
		SDS-Net               & \textbf{82.26}     & \textbf{82.45}    & \underline{89.71}    & \textbf{100}           & \textbf{6.57}           & \textbf{95.04}     & \textbf{95.04}    & {\textbf{97.46}}    & \textbf{99.37}           & \underline{5.02}           & \textbf{67.74}     & \textbf{66.95}    & {\textbf{80.56}}    & 91.92           & \textbf{10.58} \\ \hline       
	\end{tabular}
\end{table*}
\newline \indent The experimental results under a multidimensional evaluation framework demonstrate that traditional methods lag far behind deep learning models. Our SDS-Net achieves mIoU scores of 82.26\%, 95.04\%, and 67.74\% on three benchmarks. This performance is attributed its dual-branch design that effectively integrates shallow spatial and deep semantic features for precise pixel-level discrimination.
\newline \indent Table~\ref{tab2} compares SDS-Net with existing methods on IRSTD-1K, covering model size, GFLOPs, inference speed, GPU memory, and mIoU. Compared with DNA-Net, SDS-Net achieves high accuracy with notable efficiency—reducing parameters by 57.5\%, FLOPs by 47.8\%, and GPU memory by 58.5\% without relying on compression or quantization.
\newline \indent While not as lightweight as ACM, SDS-Net offers much higher accuracy with moderate resource use, suggesting that excessive compression can compromises performance. Overall, SDS-Net strikes a strong balance between accuracy and efficiency for real-time infrared small target detection under limited resources.
\newline \indent For a comprehensive evaluation of target detection performance, this work extends beyond conventional fixed-threshold metrics (IoU, nIoU, $P_d$, $F_a$) by introducing ROC analysis to assess model adaptability across thresholds. Fig.~\ref{roc} benchmarks the proposed SDS-Net against state-of-the-art methods. SDS-Net excels in ROC performance across three datasets, showing rapid TPR escalation in low-FPR regions (0--0.2) for enhanced sensitivity, while maintaining $>85\%$ accuracy in high-FPR scenarios ($>$0.5). These results confirm its operational robustness and practical viability in cluttered infrared environments.
\begin{table}[t]
	\caption{The number of model parameters(M), the number of floating-point operations(G), the inference time for each image(S) of different methods, the GPU memory usage(M),and mIoU($\%$) when the batch size is 16 on the IRSTD-1K dataset are compared.}
	\label{tab2}
	\centering
	\renewcommand{\arraystretch}{1.3}
	\setlength\tabcolsep{1.6mm}
	\begin{tabular}{cccccc}
		\hline \hline
		Method   & Params & Flops & Inference & Memory & mIoU\\ \hline
		ACM ~\cite{17}  & 0.398  & 0.402 & 0.029 & 897 & 59.32     \\
		RDIAN ~\cite{28} & 0.216  & 3.718 & 0.035 & 3331 &  65.25    \\
		ISTDU-Net ~\cite{45} & 2.751  & 7.944 & 0.039 & 10717 &  63.74    \\
		ISNet ~\cite{27} & 0.966  & 30.618 & 0.058 & 15178 &  64.36    \\
		UIU-Net ~\cite{19} & 50.540  & 54.425 & 0.060 & 14747 &  66.15    \\
		SCTransNet ~\cite{21}  & 11.19  & 20.24 & 0.067 & 7077 & 66.33   \\
		DNA-Net ~\cite{18} & 4.696  & 14.261 & 0.050 & 9869 &  65.90   \\
		SDS-Net (Ours)  & 2.701  & 6.823 & 0.048 & 5777 &  67.74    \\
		 \hline
	\end{tabular}
	\vspace{-3mm}
\end{table}

\begin{figure*}[t]
    \centering
    \includegraphics[width=.99\textwidth]{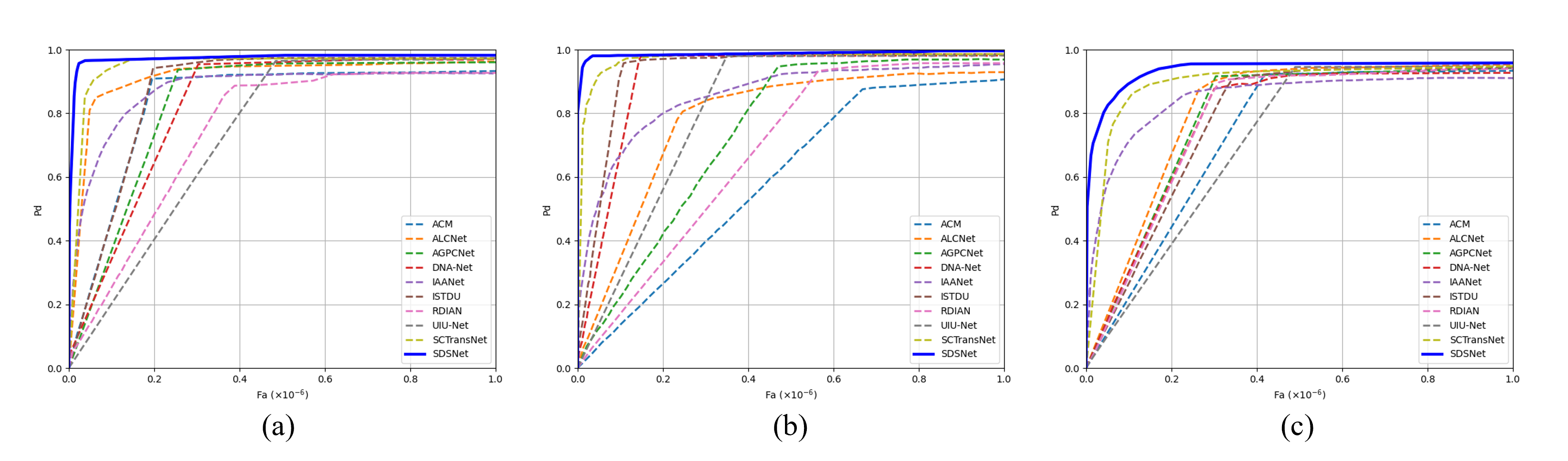}
    \caption{ROC curves of different methods on the NUAA-SIRST, NUDT-SIRST, and IRSTD-1K datasets. (a) NUAA-SIRST. (b) NUST-SIRST. (c) IRSTD-1K.}
    \label{roc}
\end{figure*}

\subsection{Ablation study}
\label{sec4-E}
%\noindent {\textbf{Datasets:}} 
This section presents ablation studies on the NUAA-SIRST dataset using a U-Net baseline. As illustrated in Table~\ref{tab3}, key modules—including RBs, DS, the shallow module, the deep module, and ADSF—are added sequentially, each yielding performance gains. Notably, the shallow module alone improves the mIoU, nIoU, and F-measure by 5.85\%, 4.66\%, and 2.29\%, respectively, underscoring the value of rich shallow feature extraction for accurate detection.
\begin{table*}[t]
	\caption{Ablation experimental results of SDS-Net on the NUAA-SIRST dataset.}
	\label{tab3}
	\centering
	\renewcommand{\arraystretch}{1.3}
	\setlength\tabcolsep{1.6mm}
	\begin{tabular}{ccccccccc}
		\hline \hline
		U-Net & +RBs & +DS & +Shallow Module & +Deep Module & +ADSF & mIoU ($\%$) & nIoU ($\%$) & F1-score($\%$) \\ \hline
		\ding{51}  & \ding{55}  & \ding{55} & \ding{55} & \ding{55} & \ding{55} & 72.45 & 73.68 & 83.25   \\
		\ding{51}  & \ding{51}  & \ding{55} & \ding{55} & \ding{55} & \ding{55} & 74.26 & 75.04 & 84.97   \\
		\ding{51}  & \ding{51}  & \ding{51} & \ding{55} & \ding{55} & \ding{55} & 75.01 & 76.59 & 85.26   \\
		\ding{51}  & \ding{51} & \ding{51} & \ding{51} & \ding{55} & \ding{55} & 80.86 & 81.25 & 87.55   \\
		\ding{51}  & \ding{51}  & \ding{51} & \ding{51} & \ding{51} & \ding{55} & 81.37 & 81.95 & 88.68   \\
		\ding{51}  & \ding{51}  & \ding{51} & \ding{51} & \ding{51} & \ding{51} & 82.26 & 82.45 & 89.71   \\
		\hline
	\end{tabular}
\end{table*}
\newline \indent Ablation studies based on SCTransNet (Table~\ref{tab4}, NUAA-SIRST) yield steady performance gains with the stepwise integration of the shallow module, deep module, and ADSF. Adding the shallow module improves the mIoU, nIoU, and F-measure by 0.72\%, 0.55\%, and 0.13\%, respectively. The Deep Module results in further gains of 0.76\%, 0.88\%, and 1.22\%, respectively. Replacing the CCA with ADSF yields additional improvements by 0.94\%, 0.78\%, and 1.04\% by modeling cross-level dependencies and adaptive fusion weights.
\newline \indent The next section compares MSCA, MDFA, and ADSF with existing IRSTD fusion strategies to validate the effectiveness of the proposed shallow–deep collaborative design in capturing local details and global semantics and to highlight each module's contribution to performance.
\begin{table*}[t]
	\caption{Ablation experimental results of SDS-Net on the NUAA-SIRST dataset.}
	\label{tab4}
	\centering
	\renewcommand{\arraystretch}{1.3}
	\setlength\tabcolsep{1.6mm}
	\begin{tabular}{ccccccc}
		\hline \hline
		SCTransNet & +Shallow Module & +Deep Module & +ADSF & mIoU ($\%$) & nIoU ($\%$) & F1-score($\%$) \\ \hline
		\ding{51}  &  \ding{55} & \ding{55} & \ding{55} & 79.84 & 80.24 & 87.32    \\
		\ding{51}  & \ding{51}  & \ding{55} & \ding{55} & 80.56 & 80.79 & 87.45    \\
		\ding{51}  &  \ding{51} & \ding{51} & \ding{55} & 81.32 & 81.67 &  88.67   \\
		\ding{51}  &  \ding{51} & \ding{51} & \ding{51} & 82.26 & 82.45 & 89.71    \\
		\hline
	\end{tabular}
\end{table*}
\newline \indent The following section provides an in-depth analysis of the proposed MSCA, MDFA, and ADSF modules, focusing on their design principles and functional roles within the overall architecture.
\newline \indent 1) The effectiveness of the proposed MSCA module in multiscale feature modeling and spatial enhancement is demonstrated through comparisons with SSCA (the standard full-level interaction module in SCTransNet) and two variants: MSCA with multihead attention (MSCA w/MH) and MSCA without multiscale mapping (MSCA w/o MSM). The results shown in Table~\ref{tab5} demonstrate that compared with our MSCA, MSCA w MH and MSCA w/o MSM exhibit reductions by 1.70\%, 1.66\%, 2.26\% and 0.94\%, 0.78\%, 1.04\% in mIoU, nIoU, and F-measure, respectively, on the NUAA-SIRST dataset. This occurs because the multihead strategy complicates the feature mapping space for infrared small targets, hindering information extraction from feature-limited objects. Therefore, in SDS-Net, we utilize single-head attention for the IRSTD. In contrast, multiscale mapping effectively captures target-specific details and potential spatial correlations in deep contextual backgrounds. In summary, MSCA minimizes missed detections while increasing the confidence of the detection process and the advantage of fine-grained representation.
\begin{table}[t]
	\centering
	\caption{Comparative experiments of MSCA with $mIoU(\%)$/$nIoU (\%)$/$F1-scores(\%)$ of other modules}
	\label{tab5}
	\renewcommand{\arraystretch}{1.3}
	\setlength\tabcolsep{1.1mm}
	\begin{tabular}{cccc}    \hline\hline
		\multirow{2}{*}{Model}    & \multicolumn{3}{c}{Datasets} \\  \cline{2-4} 
		& NUAA-SIRST            & NUDT-SIRST        & IRSTD-1K     \\ \hline
		SSCA~\cite{21}   & 79.84/80.24/87.32      & 94.33/94.12/96.95     & 66.33/66.42/79.84  \\ 
		MSCA w MH     & 80.42/80.84/88.01       & 94.17/94.32/97.12     & 66.45/66.69/79.59  \\
		MSCA w/o MSM     &  80.19/81.12/88.54      & 94.38/94.69/97.15     & 66.45/66.69/79.59  \\
		\textbf{MSCA}   & \textbf{82.26/82.45/89.71}       & \textbf{95.04/95.04/97.46}     & \textbf{67.74/66.95/80.56}  \\
		\hline
	\end{tabular}
	%\vspace{-7mm}
\end{table}
\newline \indent 2) The effectiveness of the proposed MDFA module is validated through comparisons with the CFN ~\cite{21}, CBAM ~\cite{40}, and an ablated variant (MDFA w/o PAM) that removes position-aware attention. As shown in Table~\ref{tab6}, compared with the CBAM baseline method, the MDFA w/o PAM achieves overall improvements in comprehensive evaluation metrics across two complex datasets by optimizing inter-module connection structures. With the further integration of PAM to construct the complete MDFA framework, the object detection performance is significantly enhanced: For key metrics in low-contrast scenarios and small object detection, the two datasets achieve mIoU improvements by 1.66\%/1.19\%, nIoU gains by 0.33\%/0.36\%, and F1-score increases of 0.07\%/0.64\%, respectively. Experiments verify that the position-aware module effectively improves the small object localization capability and enhances feature discriminability in low-contrast scenarios through reinforced spatial information modeling.
\begin{table}[t]
	\centering
	\caption{Comparative experiments of MDFA with Params(M)/Flops(G)/$mIoU(\%)$/$nIoU (\%)$/$F1-scores(\%)$ of other modules on the NUAA-SIRST and IRSTD-1K datasets.}
	\label{tab6}
	\renewcommand{\arraystretch}{1.3}
	\setlength\tabcolsep{1.3mm}
	\begin{tabular}{ccccc}    \hline\hline
		\multirow{2}{*}{Model}   &\multirow{2}{*}{Params}  &\multirow{2}{*}{Flops} & \multicolumn{2}{c}{Dataset} \\  \cline{4-5} 
		&           &                 & NUAA-SIRST            & NUDT-SIRST        \\ \hline
		CFN~\cite{21}  & 3.849  & 7.934     & 79.84/80.24/87.32    & 66.33/66.42/79.84   \\ 
		CBAM~\cite{40} & \textbf{2.498}  & \textbf{6.693}     & 79.65/79.97/87.46    & 66.48/66.46/79.88   \\
		MDFA w/o PAM & 2.532  & 6.757    & 80.52/80.96/87.41    & 66.55/66.59/79.72   \\
		\textbf{MDFA} & 2.701  & 6.823    & \textbf{82.26/82.45/89.71}    & \textbf{67.74/66.95/80.56}   \\
		\hline
	\end{tabular}
	\vspace{-3mm}
\end{table}
\newline \indent 3) The ADSF module was compared with the CCA, CAB, and their ablation variants without learnable factors (ADSF w/o LF).  As shown in Table~\ref{tab7}, compared with the former two mainstream feature fusion methods, ADSF w/o LF utilizes cross-correlation operations to establish cross-layer feature correlations, achieving multi-granularity cross-layer dependency modeling with fewer parameters and lower computational complexity. This enables the model to optimize efficiency while moderately improving core metrics. After further introduction of learnable factors, the model achieves significant performance improvements on the NUDT-SIRST dataset, with the mIoU increasing by 0.91\%, the nIoU growing by 0.57\%, and the F1-score increasing by 0.31\%. ADSF demonstrates that the dynamic regulatory mechanism effectively balances fusion weights between shallow spatial details and deep semantic information, thereby achieving optimal detection accuracy.
\begin{table}[t]
	\caption{Comparative experiments of ADSF with mIoU (\%)/nIoU (\%)/F1-scores (\%)/Params(M)/Flops(G) of other modules on the NUDT-SIRST dataset.}
	\label{tab7}
	\centering
	\renewcommand{\arraystretch}{1.3}
	\setlength\tabcolsep{1.6mm}
	\begin{tabular}{cccccc}
		\hline \hline
		Method   & mIoU & nIoU & F1-score & Params & Flops \\ \hline
		CCA~\cite{21}  & 94.32  & 94.12 & 96.95 & 3.268 &  8.447    \\
		CAB~\cite{47}  &  94.55 & 93.81 & 96.29 & 3.692 &  10.265    \\
		ADSF w/o LF  & 94.13  & 94.47 & 97.15 & \textbf{2.419} & \textbf{6.642}     \\
		\textbf{ADSF}  & \textbf{95.04}  & \textbf{95.04} & \textbf{97.46} & 2.701  &  6.823   \\
		\hline
	\end{tabular}
	\vspace{-3mm}
\end{table}
\newline \indent The effectiveness of the shallow and deep modules, along with the heterogeneous input strategy, is evaluated on the IRSTD-1K dataset (Table~\ref{tab8}). Introducing the shallow module yields a 66.73\% mIoU and 80.14\% F1 score with minimal overhead, outperforming the deep module alone and confirming its role in enhancing edge and texture representations. The deep module improves robustness to scale and background complexity but lacks spatial precision. Their combination achieves the best performance (67.74\% mIoU, 80.56\% F1 score), demonstrating the complementary benefits of structural detail from the shallow branch and semantic context from the deep branch. The heterogeneous input strategy further strengthens this synergy, boosting accuracy and generalization. 
\begin{table*}[t]
	\caption{In the comparison of shallow and deep architecture experiments, D and S represent deep branches and shallow branches, respectively.}
	\label{tab8}
	\centering
	\renewcommand{\arraystretch}{1.3}
	\setlength\tabcolsep{1.6mm}
	\begin{tabular}{ccccccccc}
		\hline \hline
		S & D & Params(M) & Flops(G) & mIoU($\%$) & nIoU($\%$) & F1-score($\%$) & $P_d(\%)$ & $F_a({10}^{-6})$ \\ \hline
		\ding{55}  & \ding{51}  & 1.357 & 6.293 & 63.53 & 63.47 & 73.49 & 87.61 & 23.64   \\
		\ding{51}  &  \ding{55} & 2.008 & 6.742 & 66.73 & 66.32 & 80.14 & 91.47 &  15.18  \\
		\ding{51}  & \ding{51}  & 2.490 & 6.788 & 67.74 & 66.95 & 80.56 & 91.92 & 10.58   \\
		\hline
	\end{tabular}
\end{table*}
\newline \indent We conducted ablation studies on the impact of varying shallow/deep module configurations on model performance and complexity (Table~\ref{tab9}). As the number of shallow modules increases (from 1 to 3), the model’s parameters and performance, particularly the mIoU and F1 score, improve, highlighting the importance of shallow layers in preserving edge and local features.
\newline \indent The optimal configuration (3 shallow layers, 1 deep layer) achieves peak performance (mIoU: 67.74\%, F1: 80.56\%) while maintaining acceptable computational cost. However, increasing the number of deep layers (from 1 to 3) leads to greater computational overhead and accuracy degradation because of the loss of target details from excessive downsampling.
These results demonstrate that a balanced shallow-deep configuration enhances both model performance and efficiency, confirming the effectiveness of synergistic multilevel feature learning.
\begin{table*}[t]
	\caption{Performance comparison under different settings of the number of shallow and deep layers (based on the NUAA-SIRST dataset)}
	\label{tab9}
	\centering
	\renewcommand{\arraystretch}{1.3}
	\setlength\tabcolsep{1.6mm}
	\begin{tabular}{ccccccc}
		\hline \hline
		Shallow-layer Number  & Deep-layer Number & Params(M) & Flops(G) & mIoU($\%$) & nIoU($\%$) & F1-score($\%$)   \\ \hline
		1  & 1  & 0.658 & 3.006 & 76.26 & 76.67 & 85.57    \\
		2  & 1  & 1.077 & 4.918 & 79.59 & 80.42 & 87.46   \\
		\textbf{3}  & \textbf{1}  & 2.701 & 6.823 & 82.26 & 82.45 & 89.71    \\
		3  & 2  & 7.844 & 18.424 & 81.76 & 81.88 & 89.13   \\
		3  & 3  & 22.558 & 42.389 & 81.56 & 81.71 & 88.94   \\
		\hline
	\end{tabular}
\end{table*}

\subsection{Visual results}
The visual comparison with seven state-of-the-art infrared small target detection methods on the NUAA-SIRST, NUDT-SIRST, and IRSTD-1K datasets demonstrates the superior performance of our method in complex backgrounds (Fig.~\ref{visiual} and Fig.~\ref{sailmaps}). This advantage is attributed to the dual-branch architecture of our shallow-deep synergistic detection network, which enables targeted processing of multilevel features. Compared with other algorithms, it achieves high segmentation accuracy while maintaining efficient inference speed.
\newline \indent As shown in Fig.~\ref{visiual}(2), our method successfully segments two closely positioned targets, while other deep learning approaches either fail to detect the targets or produce fragmented results. This limitation is due to their overreliance on deep semantic features, where spatial details are lost through repeated downsampling. Furthermore, as shown in Fig.~\ref{visiual}(3) and Fig.~\ref{visiual}(6), only our method and DNA-Net avoid false alarms. By effectively modeling the heterogeneous relationships between shallow and deep features, our approach enables the network to accurately distinguish true targets from background noise, even in the presence of strong interference, thereby reducing false detections.
\newline \indent The 3-D visualization of the saliency maps in the figure illustrate the detection performance of various methods across multiple infrared image samples. As shown in Fig.~\ref{sailmaps}(g), our proposed SDS-Net consistently produces responses that closely align with the ground truth (Fig.~\ref{sailmaps}(h)). The target regions exhibit strong and highly localized activations, while background areas remain effectively suppressed. This demonstrates the model’s capability for precise small target representation and robust background suppression, highlighting its effectiveness in challenging infrared scenarios.
\begin{figure*}[t]
	\centering
	\includegraphics[width=.9\textwidth]{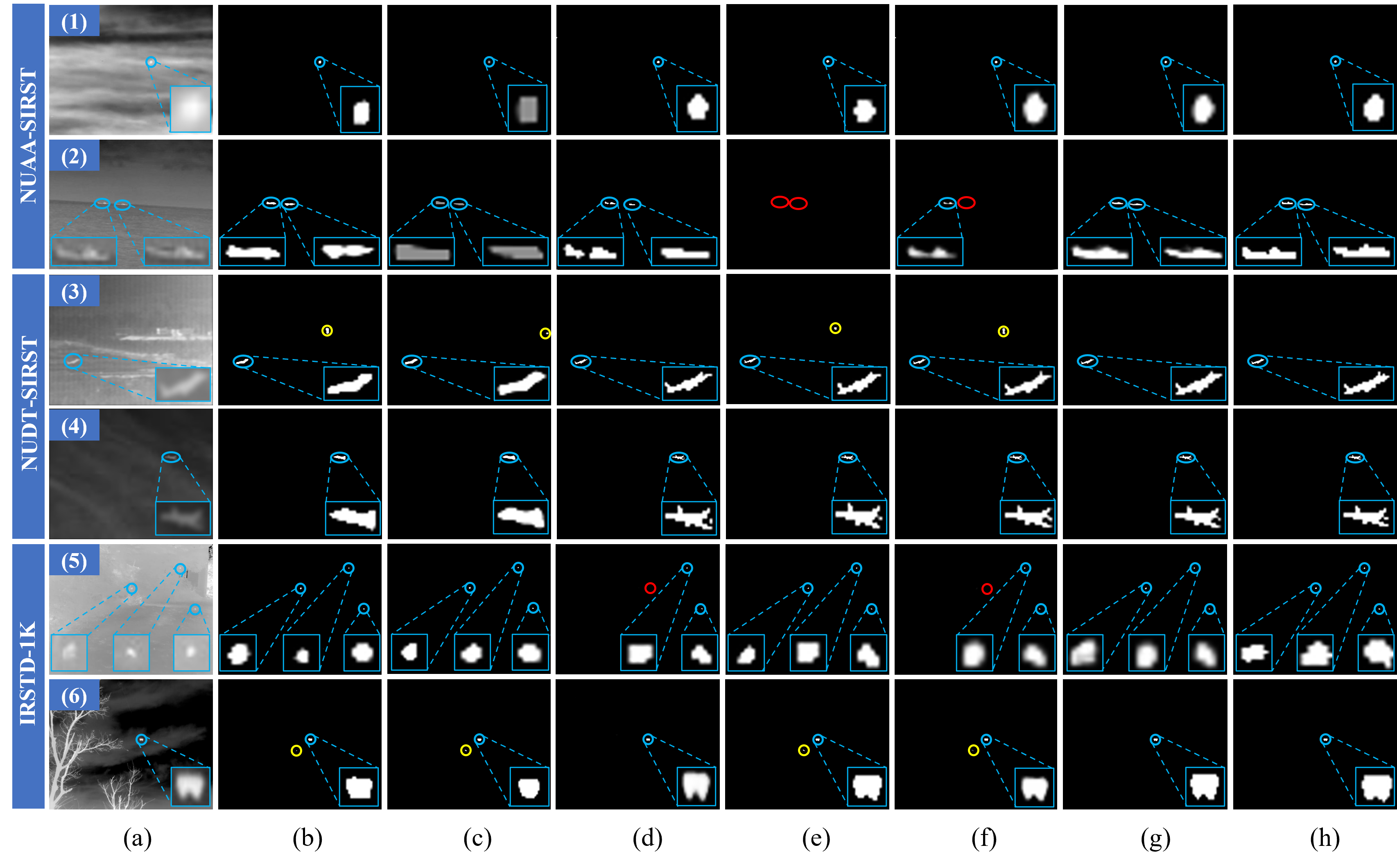}
	\caption{Visual results obtained by different IRSTD methods on the NUAA-SIRST, NUDT-SIRST, and IRSTD-1K datasets. The circles in blue, red, and yellow represent correctly detected targets, missed detections, and false alarms, respectively. (a) Input. (b) ACM. (c) ALCNet. (d) DNA-Net. (e) UIU-Net. (f) SCTransNet. (g) SDS-Net (Ours). (h) GT.}
	\label{visiual}
\end{figure*}
\begin{figure*}[t]
	\centering
	\includegraphics[width=.9\textwidth]{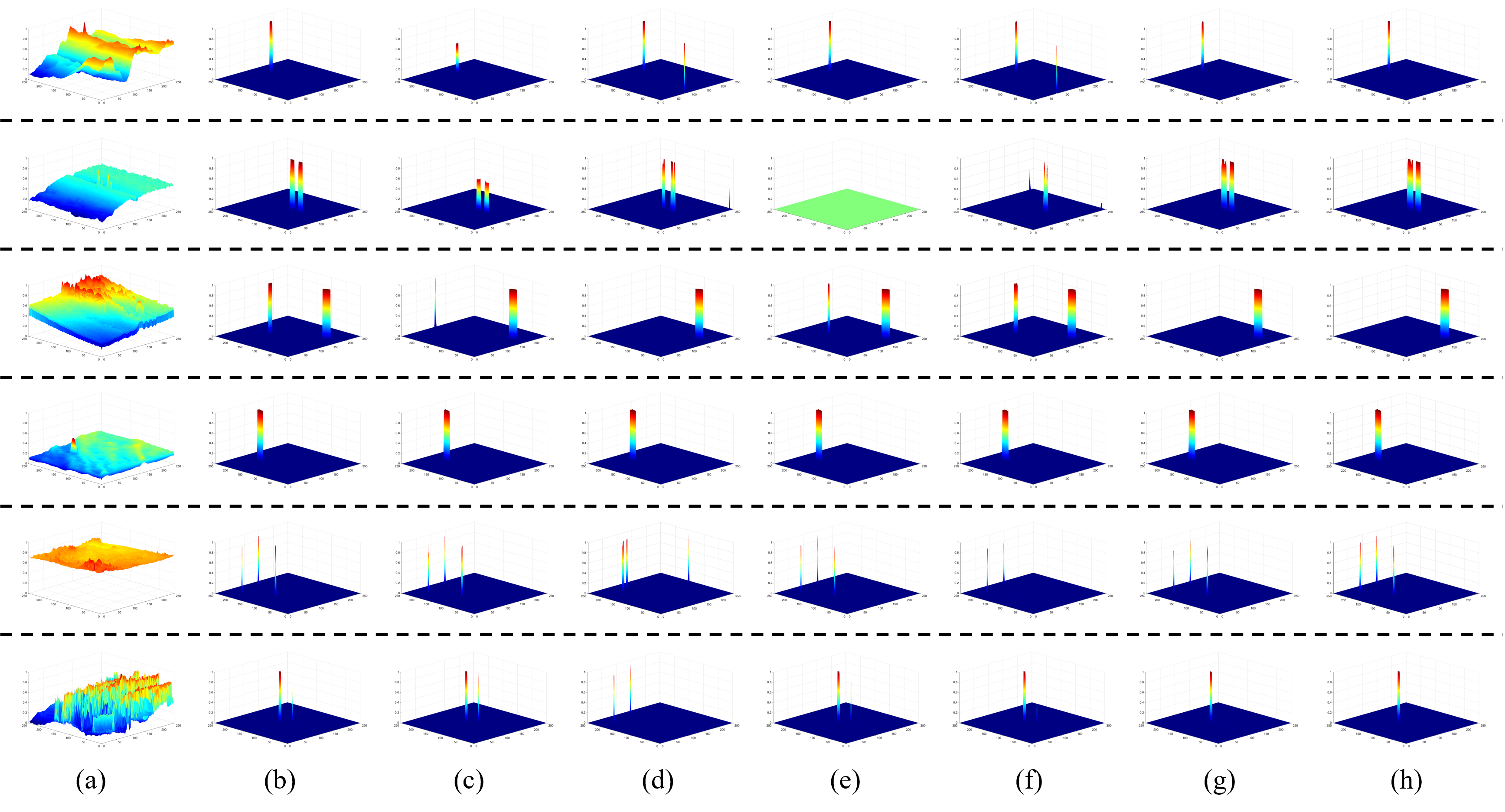}
	\caption{3-D visualization of the saliency maps of different methods on six test images. (a) Input. (b) ACM. (c) ALCNet. (d) DNA-Net. (e) UIU-Net. (f) SCTransNet. (g) SDS-Net (Ours). (h) GT.}
	\label{sailmaps}
\end{figure*}

\section{Conclusion}
\label{sec5}
This paper proposes the SDS-Net to address key challenges in infrared small target detection, including detail loss and insufficient collaboration between shallow and deep features. The network consists of two modules: 1) a shallow module for fine-grained spatial details and 2) a deep module for high-level semantic features. An ADSF module is introduced to dynamically the model interdependencies between hierarchical features, enabling effective complementary optimization. The experimental results on three benchmark datasets (NUAA-SIRST, NUDT-SIRST, and IRSTD-1K) show that SDS-Net outperforms current state-of-the-art methods in terms of detection performance while maintaining computational efficiency.
\newline \indent This study demonstrates the complementary and synergistic potential of shallow detail features and deep semantic representations in infrared small target detection. It highlights the importance of heterogeneous modeling of feature differences across layers and emphasizes the role of cross-layer collaborative mechanisms in effectively leveraging multi-level feature interactions. These insights provide valuable guidance for developing more robust and accurate detection algorithms, and lay the groundwork for future architectural designs that better exploit the hierarchical and heterogeneous nature of deep neural networks.

\bibliographystyle{IEEEtran}
\small\bibliography{IEEEabrv, reference}

\vfill

\end{document}